\documentclass[runningheads,a4paper]{llncs}
\usepackage{copyright}

\usepackage{graphicx}
\usepackage{subcaption}

\usepackage[labelformat=simple]{subcaption}

\usepackage[hidelinks]{hyperref}

\usepackage{cleveref}
\crefname{table}{Tab.}{Tabs.}
\crefname{figure}{Fig.}{Figs.}
\crefname{section}{Sec.}{Secs.}
\crefname{equation}{Eq.}{Eqs.}

\usepackage[binary-units]{siunitx}
\usepackage{pifont}

\begin{document}

\mainmatter
\title{
\textit{Robot-Relay}:
Building-Wide, Calibration-Less Visual Servoing with Learned Sensor \\ Handover Networks
}
\titlerunning{
\textit{Robot-Relay}
}
\author{
Luke Robinson, Matthew Gadd, Paul Newman, Daniele De Martini\thanks{
Thanks to EPSRC Programme Grant ``From Sensing to Collaboration'' (EP/V000748/1).
}
}
\authorrunning{ L. Robinson \textit{et al.} }
\institute{
Mobile Robotics Group, University of Oxford\\[5pt]
\texttt{lrobinson@robots.ox.ac.uk}
}
\maketitle

\copyrightnotice


\begin{abstract}
We present a system which grows and manages a network of remote viewpoints during the natural installation cycle for a newly installed camera network or a newly deployed robot fleet.
No explicit notion of camera position or orientation is required, neither global -- i.e. relative to a building plan -- nor local -- i.e. relative to an interesting point in a room.
Furthermore, no metric relationship between viewpoints is required.
Instead, we leverage our prior work in effective remote control without extrinsic or intrinsic calibration and extend it to the multi-camera setting.
In this, we memorise, from simultaneous robot detections in the tracker thread, soft pixel-wise topological connections between viewpoints.
We demonstrate our system with repeated autonomous traversals of workspaces connected by a network of six cameras across a productive office environment.
\end{abstract}

\section{Introduction}
\label{sec:introduction}


Off-board control of mobile robots from remote viewpoints holds great promise for scalable, universal autonomy -- removing the requirement for custom, expensive onboard sensors and bringing the vast cloud resource to bear on the computational problem of sensing and action.
However, the practical realisation of this problem setting requires the robot to cover large operational areas, which might not be possible with a single camera.
For this reason, multiple cameras must form a network with overlapping fields of view to reach all parts of agents' productive space.
Deployment of such a system can be time-consuming; thus, it is important to seamlessly add and remove cameras with no noticeable impact (i.e. delay or degradation) on autonomy operations.

In this work, we present a system that grows and manages a network of remote camera viewpoints with no prior knowledge of the cameras themselves.
Indeed, no explicit notion of camera position or orientation is required, neither global -- i.e. relative to a building plan -- nor local -- i.e. relative to an interesting point in a room -- nor between viewpoints.
Instead, we leverage our prior work ~\cite{robinson2023iros} in effective remote control without extrinsic or intrinsic calibration, extending it to the multi-camera setting.
In this, we memorise pixel-wise topological connections between viewpoints during the natural installation cycle for a new robot from simultaneous robot detections in the tracker thread.

We present extensive in-the-loop experiments over a network of cameras covering several rooms in a productive office space. 

\section{Related Work}
\label{sec:related_works}

Li \textit{et al.}~\cite{li2020pose} coordinate multiple cameras to improve the performance of an active tracker.
This control loop is limited to tracking objects and, therefore, camera movement, while in our work the control loop includes the remotely sensed robot.

Esterle \textit{et al.}~\cite{esterle2014socio-economic} also present a system for switching object tracking tasks between a network of cameras.
However, they too focus purely on object tracking without any control aspect of the object being tracked.

For this, D\"onmez and Kocamaz~\cite{donmez2019eye} use web cameras with lens and image plane parallel to the driving surface, while our system can handle cameras mounted at arbitrary poses.
Furthermore, in~\cite{donmez2019eye}, images from overlapping cameras are stitched together based on common features, while we learn overlap regions from demonstrations -- towards incremental, online system bring-up. 

Similarly, Elsheikh \textit{et al.}~\cite{elsheikh2016practical} explicitly solve for the relative position of overlapping cameras.
We show that this extrinsic calibration is unnecessary and opt for a topological instead of a topometric representation of the camera network.

Similar to us, Whitaker~\textit{et al.}~\cite{whitaker2020decentralised} investigate the automatic generation of waypoint graphs for each camera in an environment and path planning through the environment across multiple camera fields of view.
This, again, is based on overhead cameras facing straight down, while our system is general to perspectives in the camera network.

For industry context, consider that there are active commercial deployments such as~\cite{shalom2022systems} for the navigation of fully autonomous mobile robots through readily available ceiling-mounted cameras that maintain constant floor awareness.

\section{Technical Approach}
\label{sec:technical_approach}

\begin{figure}[!h]
\centering
\includegraphics[width=0.78\textwidth]{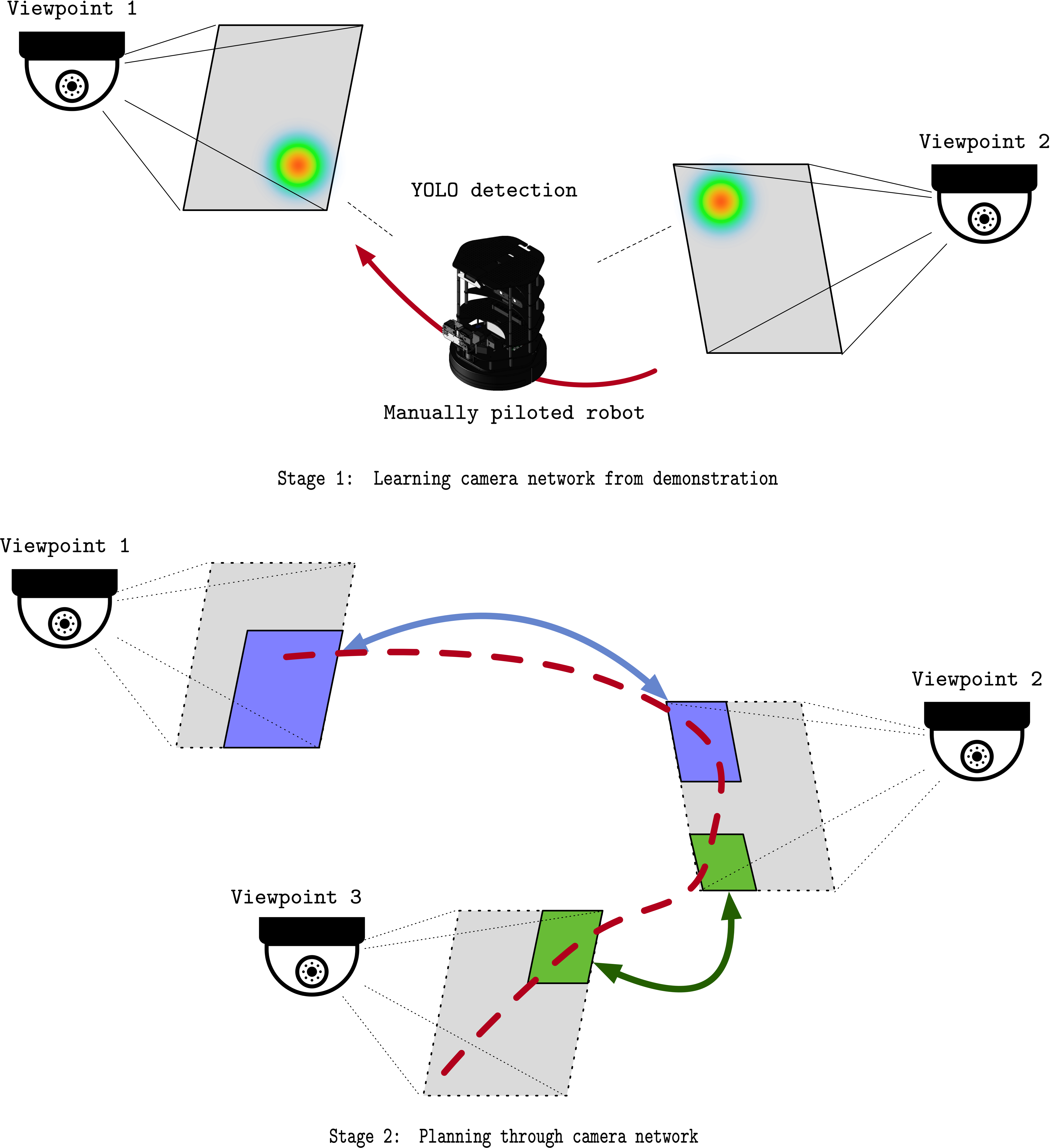}
\caption{
Collecting many simultaneous observations to learn camera overlap regions (top) enables site-wide autonomy by planning through the camera network (bottom).
}
\label{fig:inversion-iser-system}
\vspace{-25pt}
\end{figure}

Our system is illustrated in~\cref{fig:inversion-iser-system}, consisting of learning and deployment stages.

The foundational mode of autonomy, upon which our work is built, has a robot detected through a learned position and orientation methodology and controlled purely in image space through a low-level controller based on potential fields.
Based on previous work~\cite{robinson2023iros}, this preliminary system is described in~\cref{sec:single_camera}.

\Cref{sec:handover,sec:traversal} describe the \textit{core contribution of this work}, in which a graph of the known cameras is automatically constructed through incremental observations of robot traversals, and a high-level planner instructs the robot on the cameras to traverse to reach a designated goal position in a target camera.

Incidentally, for this new multi-camera setup, we have improved this base system with automatic drivable region segmentation in~\cref{sec:sam_drivable} and planning based on potential fields in~\cref{sec:llc}.

\subsection{\textit{Preliminaries}: Single-Camera Detection \& Training}
\label{sec:single_camera}

Our detection module comprises two blocks: (1) a position and (2) an orientation detector.
We directly employ the setup of our previous work \cite{robinson2023iros}, where an updated YOLOv8~\cite{terven2023comprehensive} network recovers the bounding box of the robot in operation and a learnt orientation detector \cite{robinson2023iros,ruiz2018fine} recovers the orientation in image space -- i.e. the expected direction of motion -- from these bounding boxes.

This in-camera detection system is replicated for all cameras, each with a position and orientation detector trained specifically on their individual scene through (1) labelled spins and (2) random environment walks \cite{robinson2023iros}.
For (1) the object detector is trained via \textit{labelled spins}, which involves driving the robot to five image locations and observing three \SI{360}{\degree} rotations without translation, capturing the chassis from all sides and distances.
Labels are only needed for the initial and final frames at each spin location, with interpolation for the rest, reducing manual labelling time and effort.
For (2), the orientation estimator is trained through \textit{random environment walks}, with the robot autonomously moving within a user-defined image-space boundary.
The robot randomly rotates when it hits the boundary, then continues straight until it reaches another.
For both, more detail is provided in~\cite{robinson2023iros}.

\begin{figure}
\centering
\begin{subfigure}{0.48\textwidth}
\includegraphics[width=\textwidth]{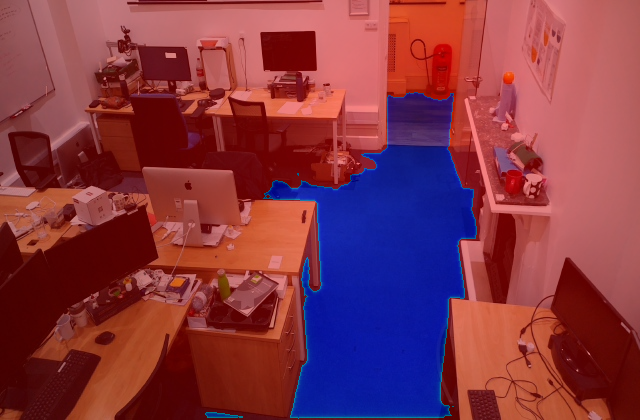}
\caption{}
\label{fig:test}
\end{subfigure}
\begin{subfigure}{0.48\textwidth}
\includegraphics[width=\textwidth]{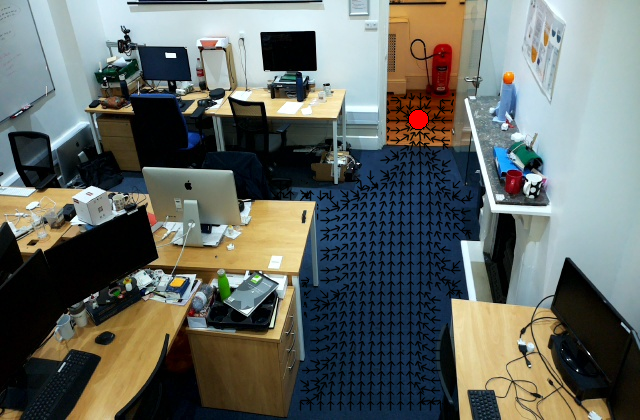}
\caption{}
\label{fig:camera1_visualisation}
\end{subfigure}
%
\caption{
\subref{fig:test} drivable region segmented in camera 7 with \textit{Segment Anything}~\cite{kirillov2023segment}.
\subref{fig:camera1_visualisation} Potential fields in camera frames, directing the robot to the handover region, where the next camera can detect and direct it to the target in the second frame. 
}
\label{fig:more_detail}
\vspace{-30pt}
\end{figure}

\subsection{Drivable Region Estimation}
\label{sec:sam_drivable}

Labelling of drivable regions can be tedious, and automatic detection can be challenging due to the huge variety of scenarios we can encounter.
Therefore, we use a semi-automatic labelling approach that exploits recent advances in foundational vision models.
In particular, the installation comprises the manual selection of a point in the drivable region through a GUI, which is passed as a prompt to \textit{Segment Anything}~\cite{kirillov2023segment}, with all other regions being considered non-drivable -- as shown in~\cref{fig:test}.
Notably, this operation is done \textit{only once per camera}.

\subsection{Low-Level Planner \& Controller}
\label{sec:llc}

With the drivable region specified, we control the robot within each camera through a potential-field method \cite{khatib1986real}, which simultaneously guides the robot towards the goal position in image space and avoids obstacles.

The potential field is calculated by combining an attractive value and repulsive value at each pixel location.
The repulsive value is calculated from the reciprocal distance of the closest obstacle, with a cutoff value determining each obstacle's region of influence.
Here, the boundaries of the drivable region are considered obstacles.
The attractive value is calculated as the square root of the distance from the pixel to the goal.
Both the attractive value and the repulsive value can be scaled using weight parameters. 

At each time step, the low-level planner finds the gradient of the potential field at the robot's current in-image location, measures the difference between this desired heading and the robot's current heading and then uses a PD controller to adjust the rotational velocity to reduce this difference while maintaining a constant linear velocity.
Examples are shown in~\cref{fig:more_detail}.

\subsection{Handover-Region Discovery and Graph Creation}
\label{sec:handover}

\begin{figure}[t]
\centering
\begin{subfigure}{0.48\textwidth}
    \includegraphics[width=\textwidth]{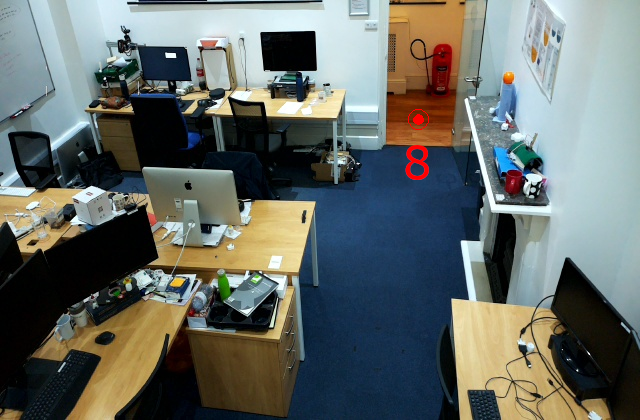}
    \caption{Camera 7}
\end{subfigure}
\begin{subfigure}{0.48\textwidth}
    \includegraphics[width=\textwidth]{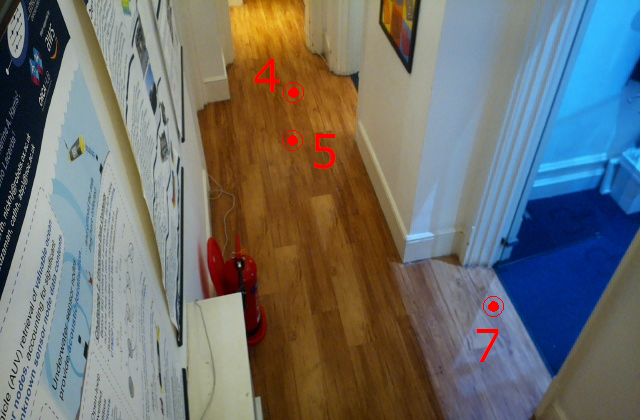}
    \caption{Camera 8}
\end{subfigure}
\caption{
Handover regions in cameras 7 (left) and 8 (right) are depicted as red markers and labelled with the relative camera ID they lead to.
For example, the red `7' in camera 8's view (right) corresponds in physical space to the red `8' in camera 7's view (left).
}
\vspace{-15pt}
\label{fig:handover}
\end{figure}

With these preliminaries descried, we now propose an automatic discovery of the handover regions among the cameras.
As described above, each camera deployment will encompass a random walk of the robot within the drivable region of the camera.
These random walks run for several minutes -- up to 30 minutes for larger fields of view.

During these random walks, the activity in each camera is monitored through the detectors and, as soon as the robot is detected in more than one camera, the topological identity of the pixels in the two cameras where the centres of the bounding boxes lie is logged.
Given a cluster of such pixel-to-pixel identities, we select the one positioned the farthest from its closest obstacle in either camera as the handover point between the two cameras and create a directed edge in a \textit{handover camera graph} between nodes representing those views.
Specifically, the handover point represents the pixel location the robot should navigate to in the current camera such that it will be seen by the other camera that the edge is pointing to.

\subsection{High-Level Planner and Graph Traversal}
\label{sec:traversal}

Through the camera handover graph, we can instruct a high-level planner to generate the cameras needed to navigate from the current robot position to a final goal destination.
To achieve this, the high-level planner is implemented as an A* algorithm, which directly operates on the camera graph, returning the shortest path from the original to the destination camera in terms of the number of handovers.

Control is then seamlessly passed from camera to camera to reach the handover region that leads to the next one in the planned path through the constellation of viewpoints.
Control is performed by the low-level controller of the camera and relinquished as soon as the robot approaches to within a certain distance of the centre of the handover region it is driving towards.
At that point, the next camera will be responsible for detecting and driving the robot towards the next handover region, or the target position if it is the final camera.

\section{Experimental Setup}
\label{sec:experimental_setup}



\begin{figure} [t]
\centering
\begin{subfigure}{0.3\textwidth}
\includegraphics[width=\textwidth]{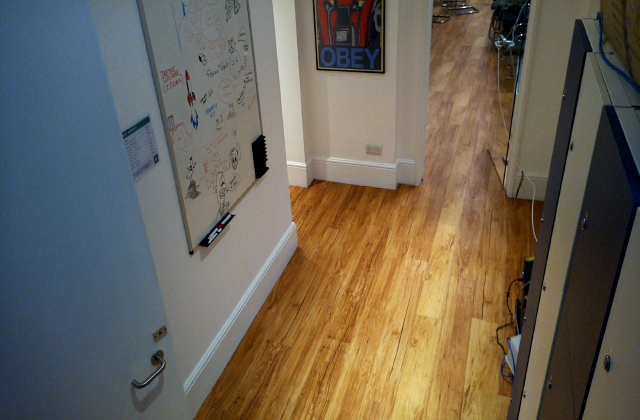}
\caption{Camera 3}
\end{subfigure}
\begin{subfigure}{0.3\textwidth}
\includegraphics[width=\textwidth]{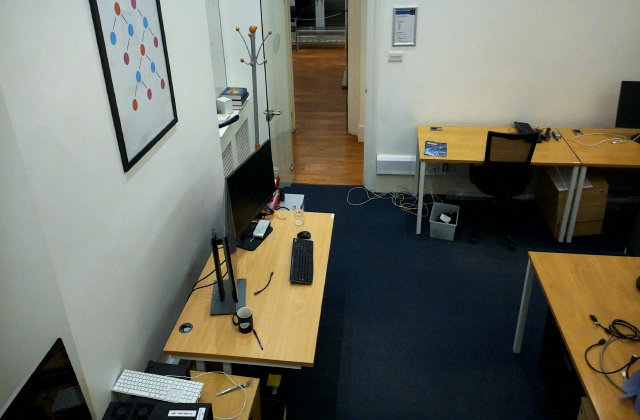}
\caption{Camera 4}
\end{subfigure}
\begin{subfigure}{0.3\textwidth}
\includegraphics[width=\textwidth]{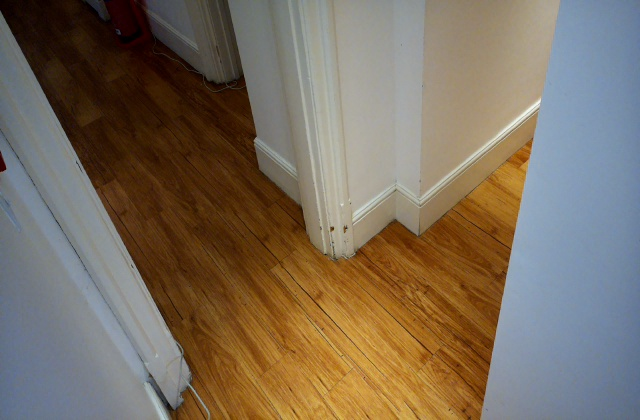}
\caption{Camera 5}
\end{subfigure}
\begin{subfigure}{0.3\textwidth}
\includegraphics[width=\textwidth]{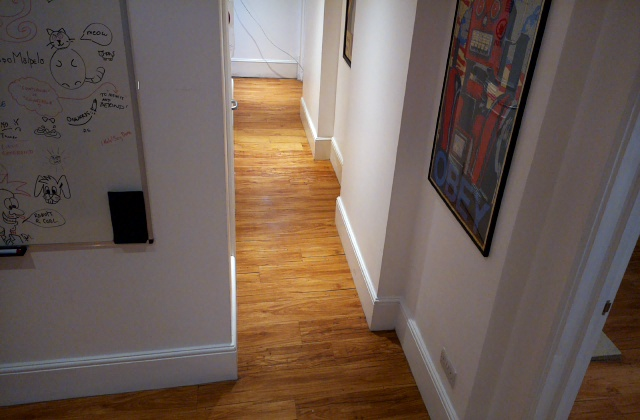}
\caption{Camera 6}
\end{subfigure}
\begin{subfigure}{0.3\textwidth}
\includegraphics[width=\textwidth]{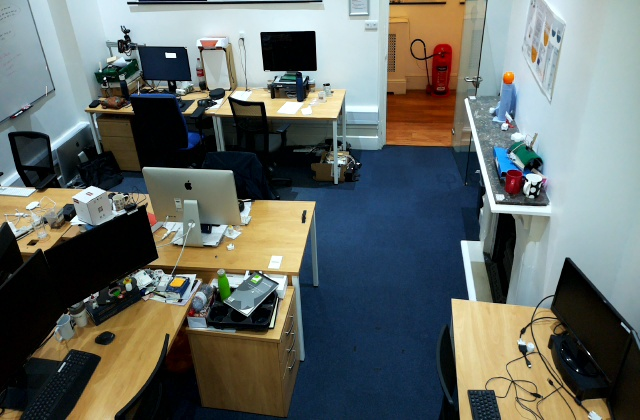}
\caption{Camera 7}
\end{subfigure}
\begin{subfigure}{0.3\textwidth}
\includegraphics[width=\textwidth]{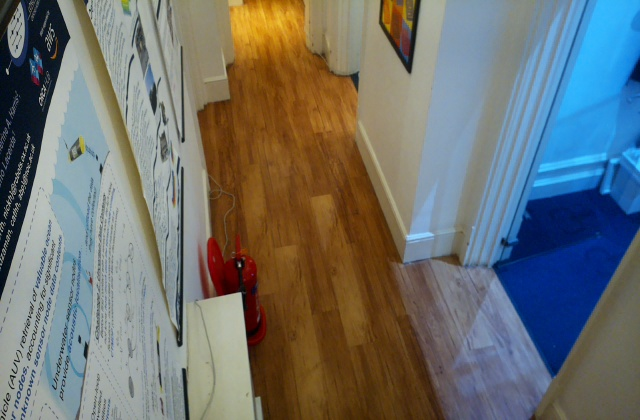}
\caption{Camera 8}
\end{subfigure}
\caption{All six viewpoints as seen by the cameras. Noticeably, they show realistic scenarios with various objects and perspective.\label{fig:camera}}
\vspace{-5pt}
\end{figure}

\begin{figure}
\centering
\includegraphics[width=\textwidth]{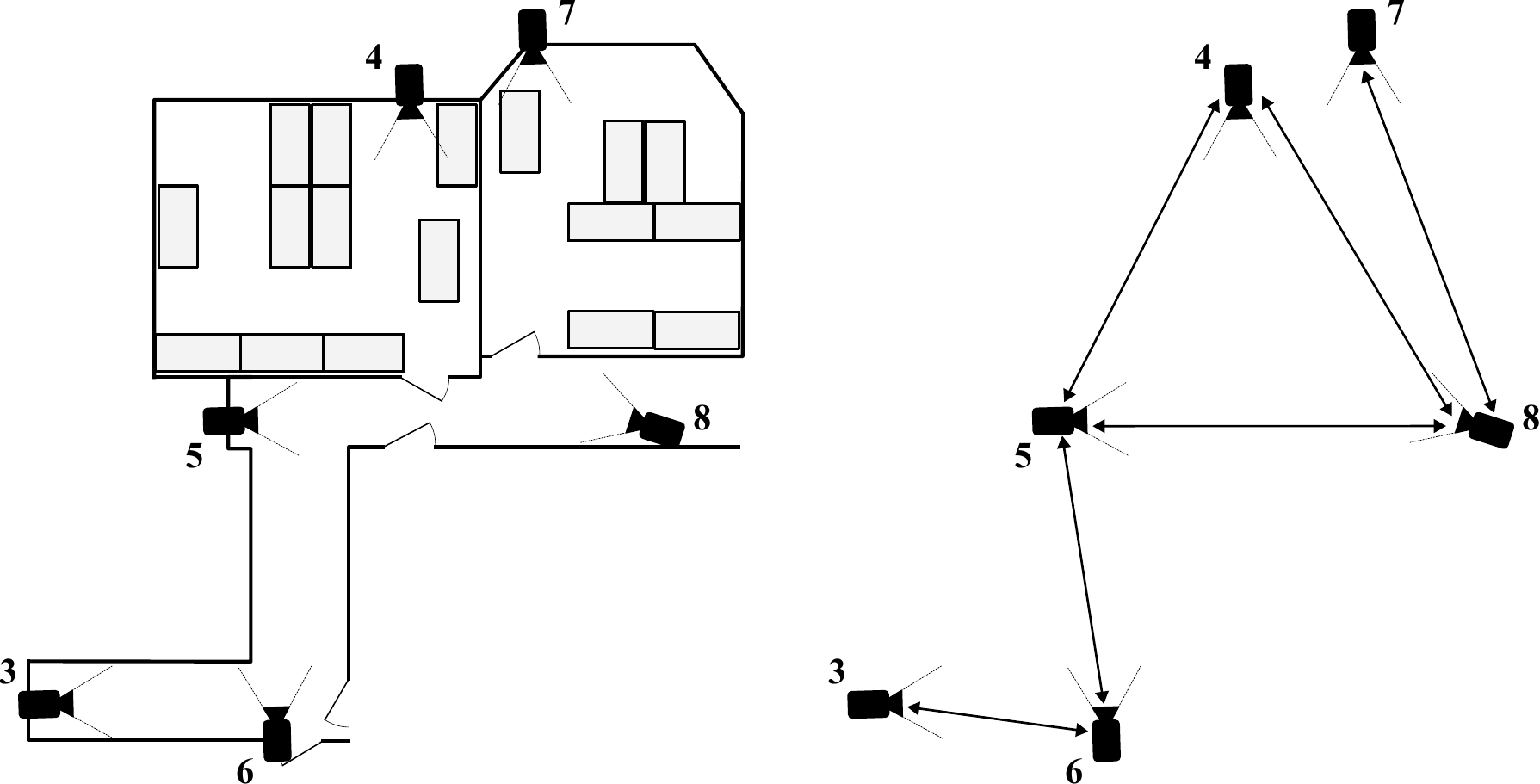}
\caption{Experimental setup (left) and camera handover graph (right). We installed six cameras in an office space, each with an overlapping field of view with at least another. These overlapping regions are shown in the handover graph, where each possible handover is depicted as an arrow.\label{fig:plan}}
\end{figure}

To test our proposed system, we deployed it in an office space, composed of three rooms and a corridor roughly \SI{13}{\metre} long.
For sufficient autonomous coverage, we set up six remote cameras throughout the building floor so that each overlaps with at least one other camera.
\Cref{fig:camera} shows the individual viewpoints for each camera, whereas the full experimental setup is depicted in \cref{fig:plan}, together with the extracted camera graph.
Notably, the rooms have different textures, clutter, object types, and dimensions, \textit{and} present different perspectives, making this deployment scenario realistic.

\begin{figure}
\centering
\begin{subfigure}{0.455\textwidth}
\includegraphics[width=\textwidth]{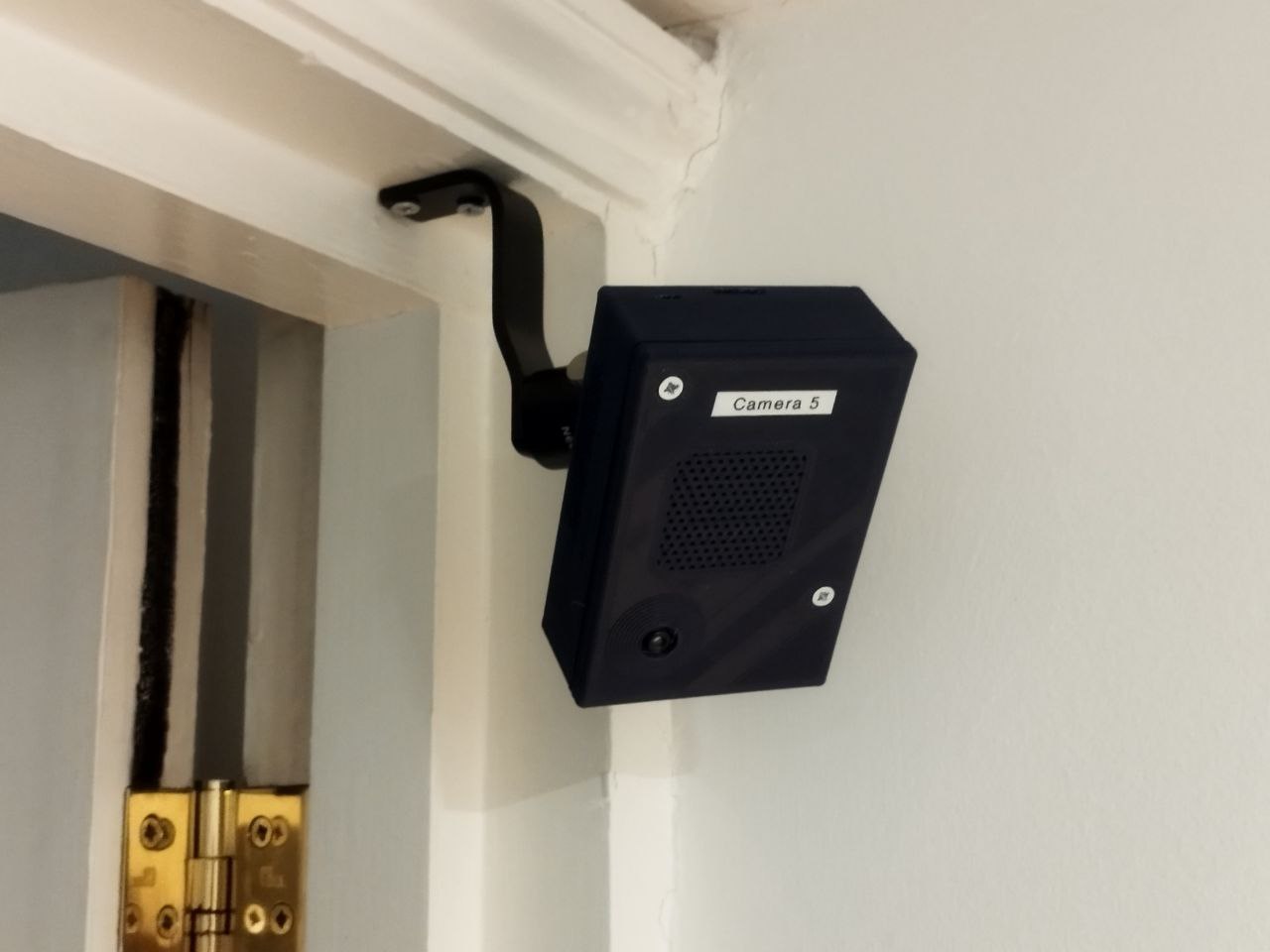}
\end{subfigure}
\begin{subfigure}{0.455\textwidth}
\includegraphics[width=\textwidth]{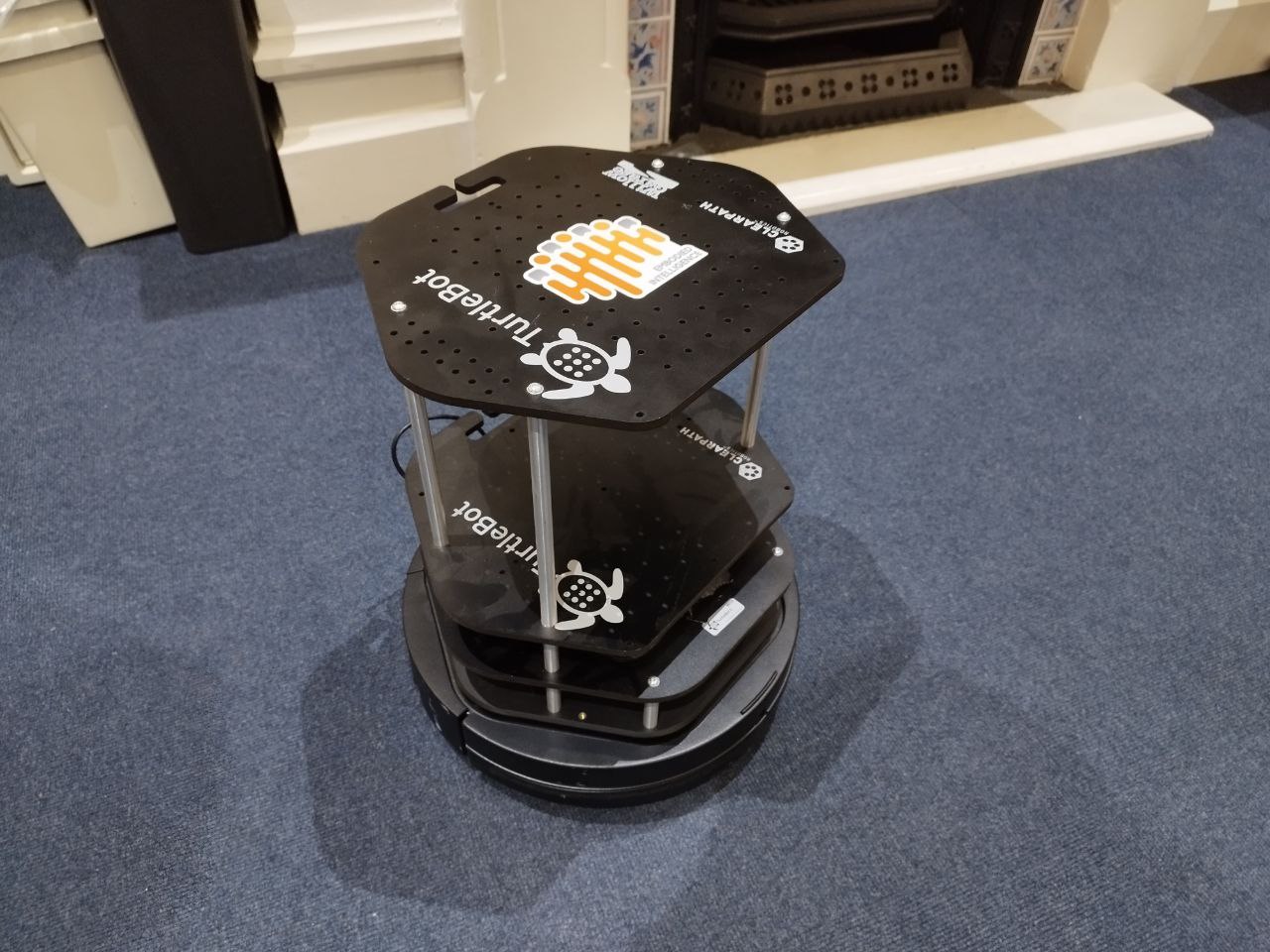}
\end{subfigure}
\caption{Hardware employed for the experiments: (left) our custom camera and (right) the Turtlebot2 platform.\label{fig:hardware}}
\vspace{-10pt}
\end{figure}

We employ for our experiments a Turtlebot2\footnote{\url{https://www.turtlebot.com}}, which we equip with a Raspberry Pi 4 Model B, used only to drive the robot through actuator commands (forward speed and rotational speed), which themselves are computed off-board and sent through WiFi.
This off-board autonomy stack runs on a server with an NVIDIA TITAN Xp and an NVIDIA GeForce GTX TITAN X.
We designed a custom camera based on a Raspberry Pi 4 Model B and a Pi Camera Module 3, connected through Ethernet to the server.
The robot and camera are shown in \cref{fig:hardware}.

We evaluate the system b7 driving amongst the cameras for five runs, which cover roughly \SI{30}{\minute}.
To do this, we generated seven random destination points covering the full graph, \texttt{\{7, 5, 3, 7, 4, 8, 6\}}, which we repeated continuously.

For the traversals, we evaluated the ability of the system to complete the path without hitting obstacles or getting stuck, as well as the time taken compared to a human, who drives the robot at the same maximum speed but looks directly at the robot in order to control its direction of travel.

For each destination point, we placed the robot at a random point in the image, marked its outline with tape and recorded its detected position in the camera.
We required that the robot to stop with its centre within this marked region to count a traversal as successful.
In practice, this was defined as the robot stopping with the centre of its detected bounding box being within five pixels of the target position.

\section{Results}
\label{sec:results}

\Cref{tab:results} shows the time taken for each traversal.
This ``traversal time'' excludes the time taken before moving to calculate a route through the camera graph and form the required potential field.
This was done over five cycles of the seven traversals described in \cref{sec:experimental_setup}, totalling roughly \SI{30}{\minute} of continuous autonomy when computation time is included.
Only three traversals were unsuccessful during this experiment, as marked with an `x' in the table.
These minor incidents are discussed in detail in \cref{sec:failure_modes}.
Combining this with the average of the times taken to complete that traversal in other runs, the system can be said to have achieved 89.6\% autonomy by time.

Further, when compared with the times taken for the same path traversals by a human, the system's efficiency can be seen -- the autonomous control was, on average, only 8.4\% slower than a human-controlled path.

\begin{table}[]
\renewcommand{\arraystretch}{1.3}
\centering
\begin{tabular}{c|c|c|c|c|c|c|c||c}
     Run  & \texttt{7$\rightarrow$5}  & \texttt{5$\rightarrow$3}  & \texttt{3$\rightarrow$7}  & \texttt{7$\rightarrow$4}  & \texttt{4$\rightarrow$8}  & \texttt{8$\rightarrow$6}  & \texttt{6$\rightarrow$7}  & Total \\
     \hline
     1 & 44.1 & 38.6 & 76.1 & 53.9 & 29.3 & 38.4 & 63.8 & 344.2  \\ 
     2 & 41.3 & 38.6 & 76.5 & 56.3 & 28.5 & 41.9 & 64.9 & 348.0  \\ 
     3 & 42.8 & 39.3 & x & 55.7 & 29.1 & 48.6 & 64.5 & 356.1  \\ 
     4 & 41.0 & 46.5 & 75.6 & 54.1 & 29.3 & 40.6 & x & 350.8  \\ 
     5 & 42.9 & 39.8 & 76.1 & 56.2 & 28.9 & x & 61.5 & 347.8  \\ 
     \hline
     Human & 38.0 & 34.9 & 73.7 & 51.8 & 27.6 & 35.9 & 60.3 & 322.2  \\ 
\end{tabular}
\vspace{5pt}
\caption{Time spent on each traversal in seconds, not including time allocated for pre-move calculations of a route within the camera graph and potential fields. Total times are calculated as the sum of the individual traversals. Failed traversals are labelled as `x' and substituted with the average of the same traversal in the successful runs. \label{tab:results}}
\vspace{-15pt}
\end{table}

\begin{figure}
\centering
\includegraphics[width=0.72\textwidth]{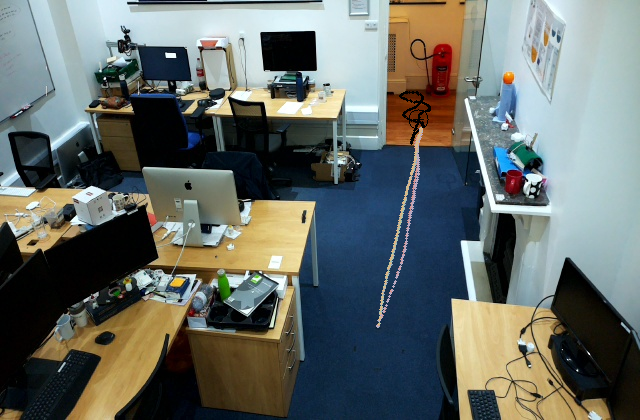}
\includegraphics[width=0.72\textwidth]{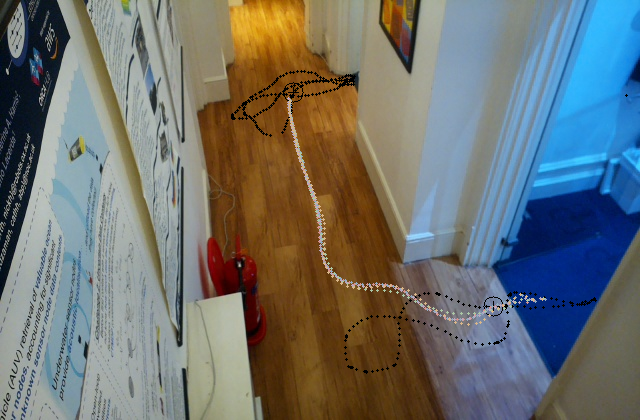}
\includegraphics[width=0.72\textwidth]{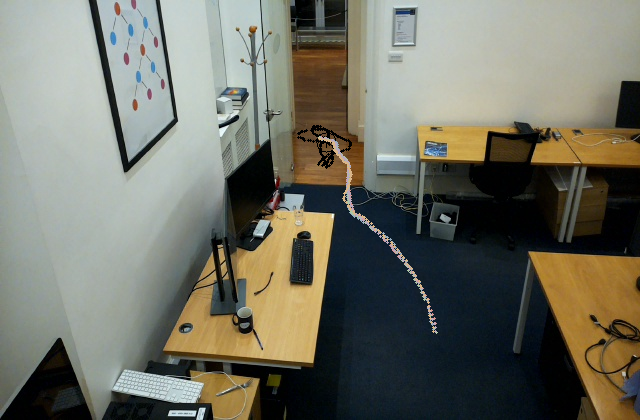}
\caption{
Multiple robot detection tracks while the vehicle was autonomously moving within and across multiple viewpoints, from many missions.
}
\label{fig:traversal}
\end{figure}

To assist in understanding how the robot performed when following paths across multiple viewpoints, one of the traversals has been shown in \cref{fig:traversal}.
Specifically, this shows all five of the traversals taken between camera 7 and camera 4, using camera 8 as an intermediate viewpoint.
Further, the black points recorded during the handover region detection phase are shown along with the selected handover point circled.
As can be seen, the robot successfully navigates to within 10 pixels of the handover point in each image, before being picked up by the subsequent camera to continue controlling it to its destination.

\subsection{Failure Modes}
\label{sec:failure_modes}

We experienced three failed traversals, with two failure modes related to the low-level potential field controller and not the graph-based camera handover, the primary focus of this work.
This demonstrates that the core principle we propose is sound, and in the future we will improve the underlying, in-image controller (using e.g. Voronoi techniques).

The first failure mode -- \texttt{3$\rightarrow$7} and \texttt{6$\rightarrow$7} failures -- made the robot indefinitely circle the goal region in the potential field of one camera but not get close enough to trigger a handover to the next camera.
In both cases, this orbiting behaviour occurred when passing through camera 5.
We hypothesise that this behaviour was caused by the area around this particular target point being outside of the influence of any obstacle.
The potential field gradient at all points around it therefore pointed radially inwards towards the target point, making the robot orbit the goal position.
To solve this issue, simple solutions could be engineered, such as checking for this condition and then reducing the linear velocity or increasing the attractive weight of the goal.
We leave the exploration of these solutions to future work.

The second failure mode -- seen in \texttt{8$\rightarrow$6} -- was caused by the robot missing the target when navigating to its final location in camera 6.
As it encountered a large heading error after it passed the target, the robot turned back around to reach the target zone eventually, but touched the wall during this manoeuvre.
For this reason, we counted the traversal as a failure.
This could most likely be resolved with further tuning of the potential field.

\subsection{Implementation Practicalities and Observations}

Through implementing our proposed system and conducting experiments, we made a number of implementation choices culminating in three main observations of the system.

A first observation was that the potential field used by the low-level controller was very sensitive to the tuning of the attractive weight towards the target region and the repulsive weight away from obstacles.
A repulsive weight that was too high might push the robot away from a handover region or create a potential well down a narrow corridor; too little and the robot would get too close to obstacles before reacting.
This meant that these parameters needed to be set for each camera individually.
Although this process required additional manual work, the static nature of the cameras meant that this need only be completed once during installation.

\begin{figure}
\centering
\includegraphics[width=0.49\textwidth]{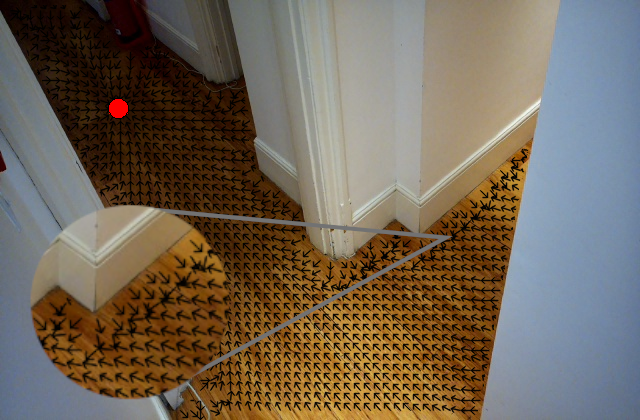}
\includegraphics[width=0.49\textwidth]{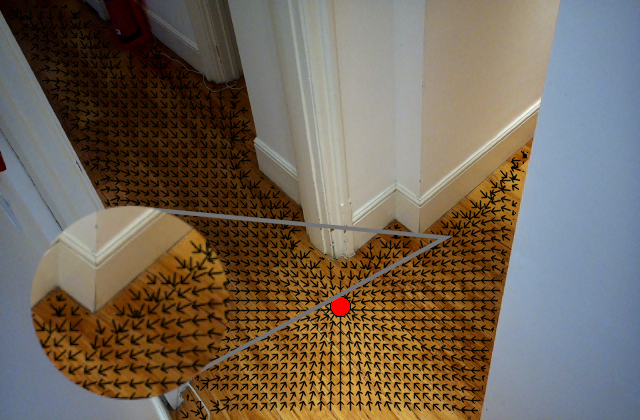}
\caption{
Potential field computed when \textit{Left} treating camera 5 as a single remote viewpoint, as compared to \textit{Right} when splitting camera 5 into two ``virtual cameras'' and placing a handover point at their interface.
}
\label{fig:5a5b}
\end{figure}

In addition to a sensitivity to tuning, certain geometries of drivable and handover regions affected the performance of the potential-field-based low-level controller.
In particular, \cref{fig:5a5b} shows how the sharp corner of the corridor, along with the placement of the original goal region on one side of the corner, caused gradients dominated by the repulsive field and gradients dominated by the attractive field to clash on the other side of the corner.
This extremely rapid change in gradient made control of the robot very impractical.
To alleviate this, during implementation the physical camera 5 was split into two ``virtual cameras'', 5a and 5b, which were connected at a handover point at the intersection of the two corridors.
Therefore, by allowing overlap between cameras 5a, 4 and 8 but not with camera 6, and similarly allowing camera 5b to overlap with camera 6 but not 4 or 8, robots traversing through camera 5 had an intermediate point to go to before they rounded the corner.
We found that this worked excellently.

One final observation was that when setting up the cameras, those that had larger overlap regions were easier to tune.
This was due to the fact that when control was passed over from one camera to another, the robot was already well within the frame of the incoming camera and therefore had room to adjust course when pushed by the potential field.



\section{Conclusion}

We have demonstrated the effectiveness of learning camera handover points and the connectivity of a network of viewpoints from demonstrations, which holds great promise for facilitating online, incremental installation, and lifelong management of off-board control from constellations of remote viewpoints.
Our experiments, involving a significant number of cameras, have showcased the system's robust autonomy, underlining its potential for real-world deployment. 

In the future, we will explore selecting amongst overlapping viewpoints depending on robot controllability -- where some typical motions are more or less observable depending on perspective.

\bibliographystyle{IEEEtran}
\bibliography{biblio}

\end{document}